\documentclass[letterpaper, 10 pt, conference]{styles/ieeeconf}
\IEEEoverridecommandlockouts 
\usepackage[bookmarks=true]{hyperref}
\usepackage{url}
\usepackage[fleqn]{amsmath}
\usepackage{float}
\usepackage{graphicx}
\usepackage{amssymb}
\usepackage{algorithm}
\usepackage{caption}
\usepackage[noend]{algpseudocode}
\usepackage{flushend}
\makeatletter
\newcommand\fs@spaceruled{\def\@fs@cfont{\bfseries}\let\@fs@capt\floatc@ruled
  \def\@fs@pre{\vspace{0.4\baselineskip}\hrule height.8pt depth0pt \kern2pt}%
  \def\@fs@post{\vspace{-0.4\baselineskip}\kern2pt\hrule\relax\vspace{-12pt}}%
  \def\@fs@mid{\kern2pt\hrule\kern2pt}%
  \let\@fs@iftopcapt\iftrue}
\makeatother

\title{\LARGE \bf AccelMPC: High-Rate, Low-Power FPGA-Accelerated\\Model Predictive Control for Tiny Drones}

\author{Andrea Grillo$^{1,2}$, Brian Plancher$^{1}$%
\thanks{This project was supported by the National Science Foundation (Award 2411369), the Toyota Research Institute, and a Burke Research Initiation Award. Any opinions, findings, conclusions, or recommendations expressed in this material are those of the authors and do not necessarily reflect those of the funding organizations. The authors thank Roy Xing and Colin Jones for their guidance and support. Contact: {\tt\footnotesize plancher@dartmouth.edu}}%
\thanks{$^{1}$ Dartmouth College $^{2}$ École Polytechnique Fédérale de Lausanne}%
}

\usepackage{siunitx}
\usepackage{booktabs}

\begin{document}
\maketitle
\thispagestyle{empty}
\pagestyle{empty}

\begin{abstract}
  Unlocking the potential of tiny aerial robots requires order of magnitude improvements in the performance of embedded edge control. In particular, although recent cached model predictive control (MPC) solvers can handle the fast system dynamics and complex constraints required for agile drone flight, their computational demands remain prohibitive for resource-constrained robots, forcing prior implementations to operate at reduced control rates. AccelMPC overcomes this challenge through an end-to-end co-design approach that jointly optimizes the solver algorithm, numerical representation, hardware mapping, and physical integration. AccelMPC pairs a co-designed FPGA-accelerated alternating direction method of multipliers (ADMM)-based MPC solver with a custom 6~g PCB, providing high-bandwidth communication for deployment on a 35~g Crazyflie. Hardware experiments demonstrate 1 kHz onboard constrained MPC with dynamic obstacles, up to 15.6$\times$ faster solve times and 195.4$\times$ improvement in energy-delay product over state-of-the-art embedded microcontroller-based solvers, all while scaling to optimization problems with over 20,000 optimization variables and a comparable number of constraints. We release our PCB design files, firmware, and FPGA solver code open source.
\end{abstract}

\section{Introduction} \label{sec:intro}
Tiny aerial robots are a promising solution for applications ranging from emergency search and rescue to routine monitoring and maintenance of infrastructure and equipment~\cite{mcguire2019minimal,duisterhof2021sniffy,zhou2022swarm}. However, despite  advances in large-scale autonomous drone control~\cite{kaufmann2023champion,hanover2024drone}, the onboard computational budgets in terms of size, weight, and power (SWaP) severely constrain the performance of such tiny systems~\cite{neuman2022tiny}. This is a particular challenge given that the fast and unstable dynamics of such systems demand feedback rates approaching the kilohertz regime for complex tasks.

Model Predictive Control (MPC)~\cite{rawlingsModelPredictiveControl2017,mayneConstrainedModelPredictive2000,camachoModelPredictiveControl2007} has been used in prior aerial robotics work to  deliver such high-performance feedback control while explicitly accounting for system dynamics and constraints, enabling constrained trajectory tracking, collision avoidance, perception-aware control, learned high-speed models, and time-optimal flight~\cite{kamel2017robust,falanga2018pampc,torrente2021data,romero2022model,romero2022replanning, krinner2024mpccpp}. Most importantly, recent advances in cached MPC, such as TinyMPC~\cite{nguyen2024tinympc}, demonstrate how real-time operation can be achieved on the limited compute platforms available onboard tiny drones. 
However, at kilohertz control rates, the alternating direction method of multipliers (ADMM)-based~\cite{boydDistributedOptimizationStatistical2011} solvers in those works can complete only one or two iterations per control step, which is insufficient for reliable convergence when constraints are active, thus requiring operation at reduced control rates~\cite{nguyen2024tinympc,mahajan2026conic,mahajan2025robust,mahajan2026tinysdp}.
Consequently, a fundamental gap remains between the predictive, explicitly constrained control promised by MPC and the capabilities available onboard highly resource-constrained robots.

AccelMPC closes this gap through end-to-end hardware-algorithm co-design, jointly optimizing solver structure, numerical representation, hardware mapping, and physical integration onboard our target tiny-drone platform. AccelMPC pairs a co-designed FPGA-accelerated ADMM-based MPC solver with a custom \SI{6}{\gram} PCB that provides high-bandwidth communication for real-time robot deployments. We realize the same solver through two complementary FPGA mappings, exposing the hardware trade-off between energy-efficient and scalable high-rate control. 

\begin{figure*}[!t]
  \centering
  \includegraphics[trim={0pt 50pt 0 20pt}, clip, width=\linewidth]{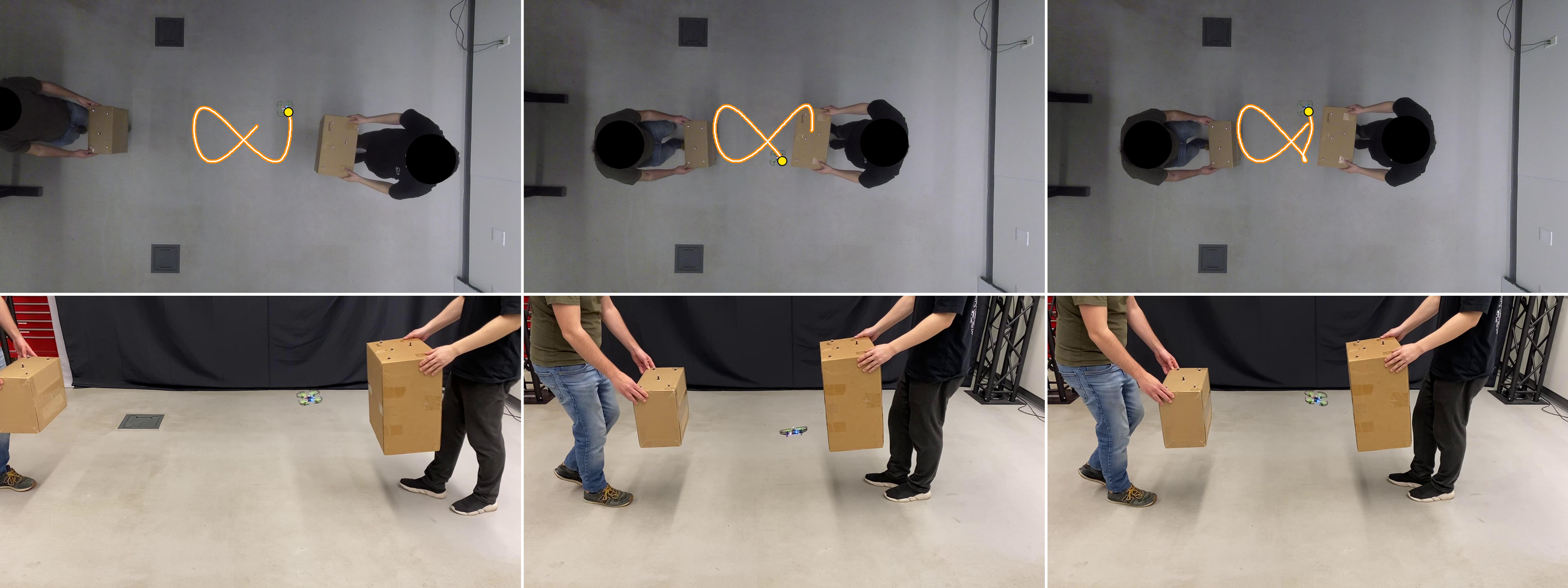}
  \caption{1 kHz onboard dynamic obstacle avoidance. From left to right: (a) nominal trajectory before interaction, (b) the quadrotor approaches a moving obstacle, and (c) the quadrotor adapts its trajectory online to safely avoid the obstacle.}
  \label{fig:dynObsAvoid}
  \vspace{-10pt}
\end{figure*}

Overall, this tight co-design enables \SI{1}{\kilo\hertz} fully onboard constrained MPC with runtime-updated bounds, demonstrated through dynamic obstacle avoidance on a \SI{35}{\gram} Crazyflie quadrotor (Fig.~\ref{fig:dynObsAvoid}). This represents up to a 15.6$\times$ faster solve time and 195.4$\times$ improvement in energy-delay product over state-of-the-art embedded solvers, while supporting more than 20,000 optimization variables and a comparable number of constraints.

Together, these results establish FPGA hardware-algorithm co-design as an enabling approach for high-bandwidth, explicitly constrained predictive control on highly compute-limited robotic platforms. 
We release our project open-source, including PCB design files, firmware, and FPGA solver code at \texttt{\url{github.com/A2R-Lab}}.

\section{Background} \label{sec:background}
\subsection{Linear MPC Formulation}
We consider the following linear MPC problem with a prediction horizon $H$, states $x_k\in\mathbb{R}^n$, and controls $u_k\in\mathbb{R}^m$, where $A$ and $B$ define the linear system dynamics, $Q \succeq 0$, $Q_f \succeq 0$, and $R \succ 0$ define the quadratic cost, and the constraint sets $\mathcal{X}$ and $\mathcal{U}$ are assumed polyhedral:
\begin{align}
  \min_{\{x_k,u_k\}} &\;\sum_{k=0}^{H-1}\!\big[(x_k{-}x_{\rm ref})^\top Q(x_k{-}x_{\rm ref}) \notag\\
               &\quad + (u_k{-}u_{\rm ref})^\top R(u_k{-}u_{\rm ref})\big] \notag\\
               &\quad + (x_H{-}x_{\rm ref})^\top Q_f(x_H{-}x_{\rm ref}) \label{eq:mpc}\\
  \text{s.t.}  &\quad x_{k+1}=A x_k + B u_k,\quad k=0,\dots,H-1,\notag\\
               &\quad x_0=\hat{x}, \quad x_k \in \mathcal{X},\quad u_k \in \mathcal{U}.\notag
\end{align}
    By stacking states and inputs into an interleaved vector,
\begin{equation}
  w = [x_0^\top,\, u_0^\top,\, x_1^\top,\, u_1^\top,\, \dots,\, x_{H-1}^\top,\, u_{H-1}^\top,\, x_H^\top]^\top,
\end{equation}
we can write~\eqref{eq:mpc} as a sparse convex quadratic program:
\begin{equation}
  \min_{w} \;\tfrac{1}{2} w^\top P w + c^\top w 
  \quad \text{s.t.}\quad l \le M w \le u,
  \label{eq:qp}
\end{equation}
where $P \succeq 0$ is block-diagonal, $M$ is sparse and block-banded due to the stage-wise dynamics coupling, and reference tracking is encoded by the linear term $c$. Equality constraints are encoded as rows of \eqref{eq:qp} with $l_i = u_i$. The interleaved ordering of $w$ keeps all coupling local to each stage, generating a banded matrix structure we exploit in Sec.~\ref{subsec:algselect}. We assume that the matrices $P$ and $M$ are fixed, while the vectors $l$ and $u$ can be updated at runtime.

\subsection{ADMM for Solving the Linear MPC QP}
\label{subsec:admm}
We review ADMM~\cite{boydDistributedOptimizationStatistical2011}, as applied to~\eqref{eq:qp}.
Introducing the auxiliary variable $y \in \mathbb{R}^m$ results in the following equivalent problem where $w \in \mathbb{R}^n$, $M \in \mathbb{R}^{m\times n}$, and $I_{\mathcal{C}}(y)$ is the indicator function of the convex set $\mathcal{C} = \{ y \mid l \le y \le u \}$:
\begin{equation} \label{eq:admm_qp}
    \min_{w,y} \;\tfrac{1}{2} w^\top P w + c^\top w + I_{\mathcal{C}}(y)
    \quad \text{s.t.}\quad Mw = y.
\end{equation}
The augmented Lagrangian of this problem is as follows where the dual variable $\lambda \in \mathbb{R}^m$ is the unscaled Lagrange multiplier for the constraint $Mw-y=0$:
\begin{equation} \label{eq:admm_augmented_lagrangian}
\begin{split}
    \mathcal{L}_\rho(w,y,\lambda) =& \tfrac{1}{2} w^\top P w + c^\top w + I_{\mathcal{C}}(y)\\
    &+ \tfrac{\rho}{2}\|Mw - y\|_2^2 + \lambda^\top (Mw - y).
    \end{split}
\end{equation}
We can then solve this via three-step ADMM iterations:
\begin{subequations}\label{eq:admm-update}
\begin{align}
w^{+} &= \arg\min_w \mathcal{L}_\rho(w,y,\lambda), \label{eq:admm-w}\\
y^{+} &= \Pi_{[l,u]}\!\left(Mw^{+} + \tfrac{1}{\rho}\lambda\right), \label{eq:admm-y}\\
\lambda^{+} &= \lambda + \rho(Mw^{+}-y^{+}), \label{eq:admm-lambda}
\end{align}
\end{subequations}
where ${}^+$ denotes the next iterate, $\Pi_{[l,u]}(\cdot)$ is the projection onto the box $[l,u]$ and the $w$-update reduces to the structured linear system solve (where $M^\top\lambda$ maps the dual variable into the primal space of the linear-system right-hand side):
\begin{equation}
\label{eq:systemSolve}
    (P + \rho M^\top M)\, w^{+}
    = -c + \rho M^\top y - M^\top \lambda.
\end{equation}
This structured solve is particularly amenable to hardware acceleration (Section~\ref{subsec:algselect}).
In our implementation, we use the scaled dual variable $s=\lambda/\rho$ and maintain $b=\rho M^\top(y-s)$, so that the linear-system right-hand side is $b-c$.

\section{Related Work} \label{sec:rel_work}
\subsection{Embedded and Edge MPC Solvers}
Given the structured QP formulation described in Sec.~\ref{sec:background}, a wide range of embedded and edge MPC solvers have been proposed with different trade-offs~\cite{ferreauEmbeddedOptimizationMethods2017}. Popular approaches include code-generation tools such as CVXGEN~\cite{mattingleyCVXGENCodeGenerator2012} which produce solvers tailored to a specific QP structure, enabling efficient execution on edge CPUs. More general-purpose approaches, such as OSQP~\cite{stellatoOSQPOperatorSplitting2020}, rely on operator-splitting methods with warm-starting and have been widely adopted in robotics. Structured solvers and frameworks such as HPIPM~\cite{frisonHPIPM2020} and acados~\cite{Verschueren2021} further exploit problem structure and optimized linear algebra to achieve high performance on embedded platforms, while robotics-oriented solvers such as ALTRO-C target fast conic MPC on more capable onboard computers~\cite{altroc}. Another thread has investigated GPU-accelerated MPC, exploiting data-level parallelism for high-performance~\cite{schubigerGPUAccelerationADMM2020,adabagMPCGPURealTimeNonlinear2024a,amatucci2025primal}. 

Tiny quadrotors like the Crazyflie, however, can support neither power-hungry edge CPUs nor GPUs, and instead rely on resource-constrained microcontrollers (MCUs).
Recent advances in cached MPC, such as TinyMPC~\cite{nguyen2024tinympc}, provide the first glimpses into the power of real-time MPC on-board tiny platforms like the Crazyflie~\cite{giernacki2017crazyflie}. These approaches leverage ADMM to reformulate the primal update as an unconstrained LQR problem and cache the associated Riccati factorizations offline, enabling efficient QP solves within tight memory and timing constraints. 
However, when constraints become active and the number of ADMM iterations required for convergence grows, even these approaches are unable to scale on MCUs, requiring operation at reduced control rates.

\subsection{FPGA-Based Linear MPC}

A significant body of work has explored FPGA acceleration of linear MPC~\cite{mcinerneySurveyImplementationLinear2018}:
Jerez \textit{et al.}~\cite{jerez2013embedded,jerezEmbeddedOnlineOptimization2014} demonstrated fixed-point first-order methods and sparse QP solvers at very high rates; 
Hartley \textit{et al.}~\cite{hartleyPredictiveControlUsing2014} targeted aircraft control with FPGA-in-the-loop simulation; 
Shukla \textit{et al.}~\cite{shuklaSoftwareHardwareCode2017} proposed a code-generation framework for FPGA splitting methods; and
Jeong \textit{et al.}~\cite{jeongWhenFPGAsMeet2023} implemented ADMM for power electronics.
To the best of our knowledge, all prior FPGA-based MPC implementations either target \emph{explicit} MPC~\cite{tondelComputationApproximationPiecewise2002}, which pre-computes control laws entirely offline, use development boards too large or power-hungry for airborne deployment, or validate only in simulation or hardware-in-the-loop.
As such, none have been deployed and flown onboard an aerial robot.
Similarly, while the neural thrust controller of Azem \textit{et al.}~\cite{azemFPGABasedNeuralThrust2024} showed that an FPGA can be flown on a Crazyflie, it does not perform online optimization.

\section{Hardware-Algorithm Codesign} \label{sec:hw_alg_codesign}
To the best of our knowledge, no prior work combines FPGA-accelerated \emph{online} MPC with onboard deployment and closed-loop flight validation on a tiny aerial robot. Our end-to-end hardware-algorithm co-designed system fills this gap and enables constrained MPC at \SI{1}{\kilo\hertz} rates. 

Achieving this requires satisfying the coupled control, optimization, compute, and physical-integration requirements summarized in Table~\ref{tab:codesign_requirements}.
High-bandwidth flight requires predictable kilohertz-rate feedback, while constrained MPC requires sufficient solver iterations and runtime-updatable constraints within each control cycle. FPGA acceleration is well suited to this setting because it provides programmable logic and arithmetic resources that can be organized into deterministic, deeply pipelined datapaths for fast and low-power computation without requiring chip fabrication. However, embedded FPGAs provide limited on-chip memory, and efficient implementations typically favor regular dataflow and fixed-point arithmetic over general-purpose floating-point computation. The resulting solver must therefore combine regular computation, compact storage, reduced-precision robustness, and efficient runtime adaptation. 

AccelMPC therefore jointly designs the optimization algorithm, numerical representation, hardware mapping, and physical integration around these requirements. This resulted in two custom FPGA solver mappings (Sections~\ref{subsec:algselect} and~\ref{subsec:fpga_mappings}) and a custom PCB (Section~\ref{subsec:pcb}) for deployment on a Crazyflie.
We open source the entire AccelMPC project at \texttt{\url{github.com/A2R-Lab}} including: the solver (\texttt{ADMM\_FPGA}), firmware (\texttt{crazyflie\_fpga\_firmware}), and PCB design (\texttt{Crazyflie\_FPGA\_Deck}).

\subsection{Algorithm and Numerical Co-Design}
\label{subsec:algselect}
 
At the algorithm level, the requirements in Table~\ref{tab:codesign_requirements} favor a solver with regular dataflow, deterministic execution, robustness to reduced precision, and inexpensive runtime updates. 

The ADMM algorithm presented in Sec.~\ref{subsec:admm} provides these properties through its structured computational patterns and known robustness to numerical error~\cite{boydDistributedOptimizationStatistical2011, eckstein1992douglas}. We therefore adopt the cached formulation of TinyMPC~\cite{nguyen2024tinympc}, trading a small amount of runtime flexibility for drastically improved computational performance. In particular, the matrices $P$ and $M$ and the penalty parameter $\rho$ are fixed and prepared offline, only the initial state, the reference window, and the bound vectors $l$ and $u$ change at runtime. Finally, the auxiliary and dual variables are warm started across control cycles to exploit the temporal coherence between the closely spaced problems that arise at \SI{1}{\kilo\hertz}.

\begin{table}[t] 
    \centering 
    \caption{Requirements for fast, low-power, onboard constrained MPC on tiny aerial robots.} \label{tab:codesign_requirements} 
    \begin{tabular}{@{}ll@{}} 
        \toprule 
        \textbf{Source} & \textbf{Requirement} \\ 
        \midrule 
        Robotics & Predictable \SI{1}{\milli\second} control period \\
        Constrained MPC & Sufficient online iteration budget \\
        Online operation & Runtime constraint updates \\
        Tiny aerial robot & Low power, mass, and footprint \\
        Fast, low-power & FPGA or ASIC compute \\
        Embedded FPGA & Low memory and reduced precision \\
        FPGA execution & Regular, pipeline-friendly computation \\
        Integration & Low-latency MCU-FPGA communication \\
        \bottomrule 
    \end{tabular} 
\end{table}
\begin{figure}[!t]
  \centering
  \includegraphics[width=\linewidth]{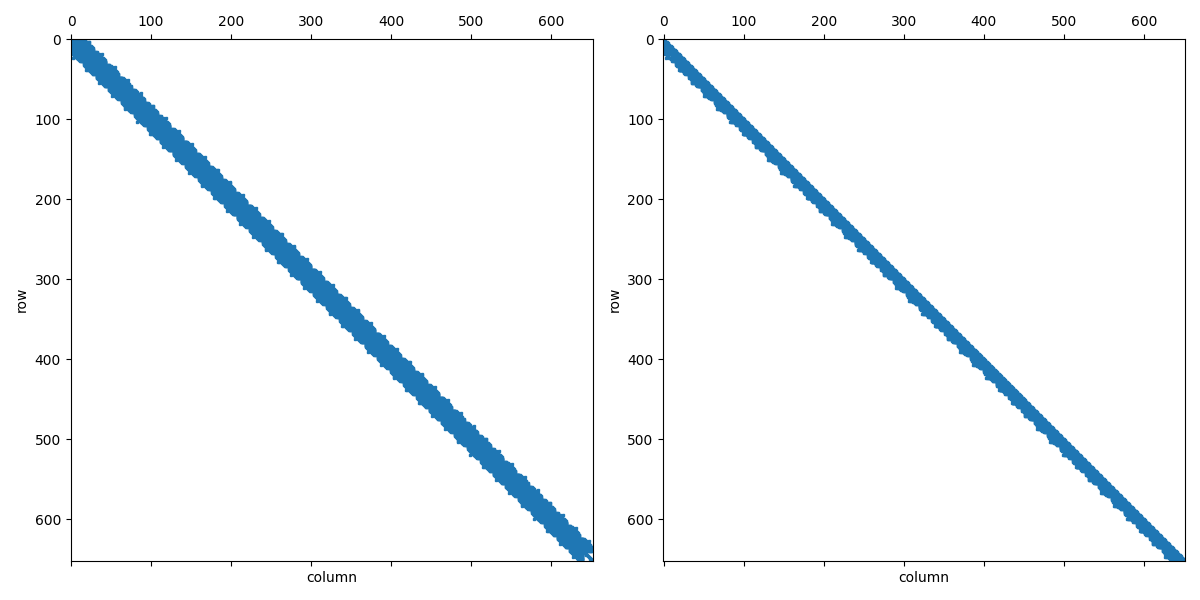}
  \caption{Left: coefficient matrix of the $w$-update. Right: corresponding
  Cholesky factor. The banded sparsity structure is preserved, enabling
  streaming forward and backward substitution with complexity scaling linearly in the
  number of optimization variables.}
  \label{fig:sparsity}
  \vspace{-15pt}
\end{figure}

Two aspects of the solver must be redesigned for efficient execution on an embedded FPGA: the structured linear solve \eqref{eq:systemSolve}, which dominates the cost of each ADMM iteration, and the numerical representation.
 
TinyMPC evaluates the structured linear solve~\eqref{eq:systemSolve} through cached Riccati-derived gain and cost-to-go matrices, executing a stage-wise backward-forward recursion of small dense matrix-vector products. This introduces a loop-carried dependency at matrix granularity, a poor match for a pipelined FPGA datapath. Prior GPU-accelerated solutions instead precompute and store the full matrix inverse~\cite{bishop2024relu}, however this exceeds the on-chip memory of an embedded FPGA.

We instead precompute a Cholesky factorization that exploits the banded structure of the coefficient matrix $P+\rho M^\top M$ (Fig.~\ref{fig:sparsity}). If the bandwidth is $h$, each row stores only $h$ values, and the online solve reduces to a forward and a backward substitution pass whose loop-carried dependencies are at scalar granularity within a window of fixed depth $h$. Since $h$ is determined by the local stage coupling and is nearly independent of the prediction horizon, the cost per ADMM iteration scales linearly with the number of optimization variables, $\mathcal{O}(N_{\rm var})$, and the substitution passes can be streamed in a hardware pipeline (Sec.~\ref{subsec:fpga_mappings}).
 
The second redesign is the numerical representation. Most critically, the solver operates in 32-bit fixed-point arithmetic, which lowers both power and latency relative to floating point on FPGA fabric. 
As our HLS implementation remains modular with respect to the numerical datatype, we validated our fixed-point implementation against an otherwise identical IEEE-754 single-precision implementation in simulation. Across the tested prediction horizons and iteration budgets, fixed-point quantization introduced negligible additional error, with differences of approximately $10^{-6}$~m or less in predicted position RMSE, $10^{-7}$~m in maximum constraint violation, and $10^{-7}$ in control RMSE.
We also restricted the penalty parameter, $\rho$, to a power of two, sacrificing some tuning freedom so that every multiplication by $\rho$ or $\rho^{-1}$ reduces to a left or right shift. Finally, the reciprocal of each diagonal element is stored in the final slot of its row of the Cholesky factor, eliminating all online division from the substitution passes. Together these choices reduce every arithmetic operation executed online to an addition, a multiplication, a shift, or a comparison, which is precisely the efficient operation set on embedded FPGAs.
 
Algorithm~\ref{alg:online-admm} summarizes the resulting online solve. We present it in the scaled-dual implementation form described above, where the right-hand side is maintained incrementally across iterations rather than reconstructed, and each line is annotated with the hardware structure that realizes it, which Sec.~\ref{subsec:fpga_mappings} describes. Finally, the solver executes a fixed number of ADMM iterations each control step to ensure deterministic latency.

\algrenewcommand\algorithmicrequire{\textbf{Offline:}}
\algrenewcommand\algorithmicensure{\textbf{Online:}}
\begin{algorithm}[t]
\caption{Co-designed fixed-iteration ADMM solve for online MPC}
\label{alg:online-admm}
\begin{algorithmic}[1]
\Require \emph{Cache compiled into the bitstream:} $M$; banded Cholesky factor, $L$, with reciprocal diagonals; $\rho$; table of reference linear terms, $c$; default nominal bounds $l,u$ 
\Ensure \emph{Dynamic Inputs:} state estimate $\hat{x}$; trajectory index; optional bound override; iteration budget $K$
\State Initial-condition $\gets \hat{x}$
\State Select $c$ for the current window \Comment{table lookup}
\State Decode and apply any runtime bound override to $(l,u)$
\State Warm start $(s, b)$ from the previous control cycle
\For{$j = 1,\ldots,K$}
  \State $r \gets b-c$ \Comment{elementwise subtraction}
  \State Solve $Lz = r$, then $L^\top w = z$ \Comment{streamed, depth $h$}
  \State $v \gets Mw$ \Comment{sparse or staged product}
  \State $y \gets \Pi_{[l,u]}(v+s)$ \Comment{compare-saturate}
  \State $s \gets s + v-y$ \Comment{elementwise add-subtract}
  \State $d \gets \rho(y-s)$ \Comment{shift}
  \State $b \gets M^\top d$ \Comment{sparse or staged product}
\EndFor
\State Store $(s,b)$ for the next control cycle
\State \Return first control input extracted from $w$
\end{algorithmic}
\end{algorithm}
 
\subsection{FPGA Datapath and Complementary Mappings}
\label{subsec:fpga_mappings}
 
Every line of Algorithm~\ref{alg:online-admm} maps to a streaming hardware structure. The substitution passes are implemented with a \emph{sliding-window shift register} of depth $h$ where newly computed values enter the window as old values shift out, so the computation operates on small local storage with fully regular access patterns that high-level-synthesis frameworks such as Vitis HLS~\cite{vitis_hls_ug1399} readily pipeline. The box projection reduces constraint handling to independent comparison-and-saturation operations, preserving a regular datapath even when constraints are active. The scaled-dual and right-hand-side updates are fused multiply-add streams over the same stage-wise structure, producing the next solve right-hand side without assembling dense matrices online.

One design axis remains open in this datapath, namely, how the constraint operators are represented. That is, they can either be stored explicitly in on-chip memory as expanded sparse operators, or the required coefficients can be regenerated from the underlying stage-wise system dynamics according to a deterministic access schedule. We implement both, as two FPGA mappings of the same solver, denoted \texttt{full\_sparse} and \texttt{staged}, respectively. Both use the same fixed-point representation, execute the same ADMM updates, and expose the same runtime interface. They differ only in operator representation, placing the dominant hardware cost at different points in the memory-computation tradeoff. As these generate tradeoffs in energy and latency (Sec.~\ref{sec:exp_ev}), no single choice is optimal across the full range of robotic MPC problem sizes or applications.
 
The \texttt{full\_sparse} mapping's direct expanded storage avoids online reconstruction and achieves the lowest measured energy per solve. However, its scalability is ultimately limited by the horizon-dependent storage of the expanded matrices, which increases BRAM demand and eventually causes spillover into distributed LUT memory.
 
The \texttt{staged} mapping instead retains a compact stage-wise representation of the MPC dynamics and reconstructs the required operator products during execution. Despite this additional reconstruction, its more streamlined data movement yields slightly lower latency at short horizons in our implementation. More importantly, its memory requirements grow substantially more slowly with problem size, allowing much larger horizons to fit on the same FPGA. This improved scalability comes at the cost of moderately higher power and energy consumption over the shared operating range.

\subsection{Custom PCB Design and Integration} \label{subsec:pcb}

\begin{figure}[!t]
  \centering
  \begin{minipage}{0.32\linewidth}
    \centering
    \includegraphics[width=\linewidth, angle=90]{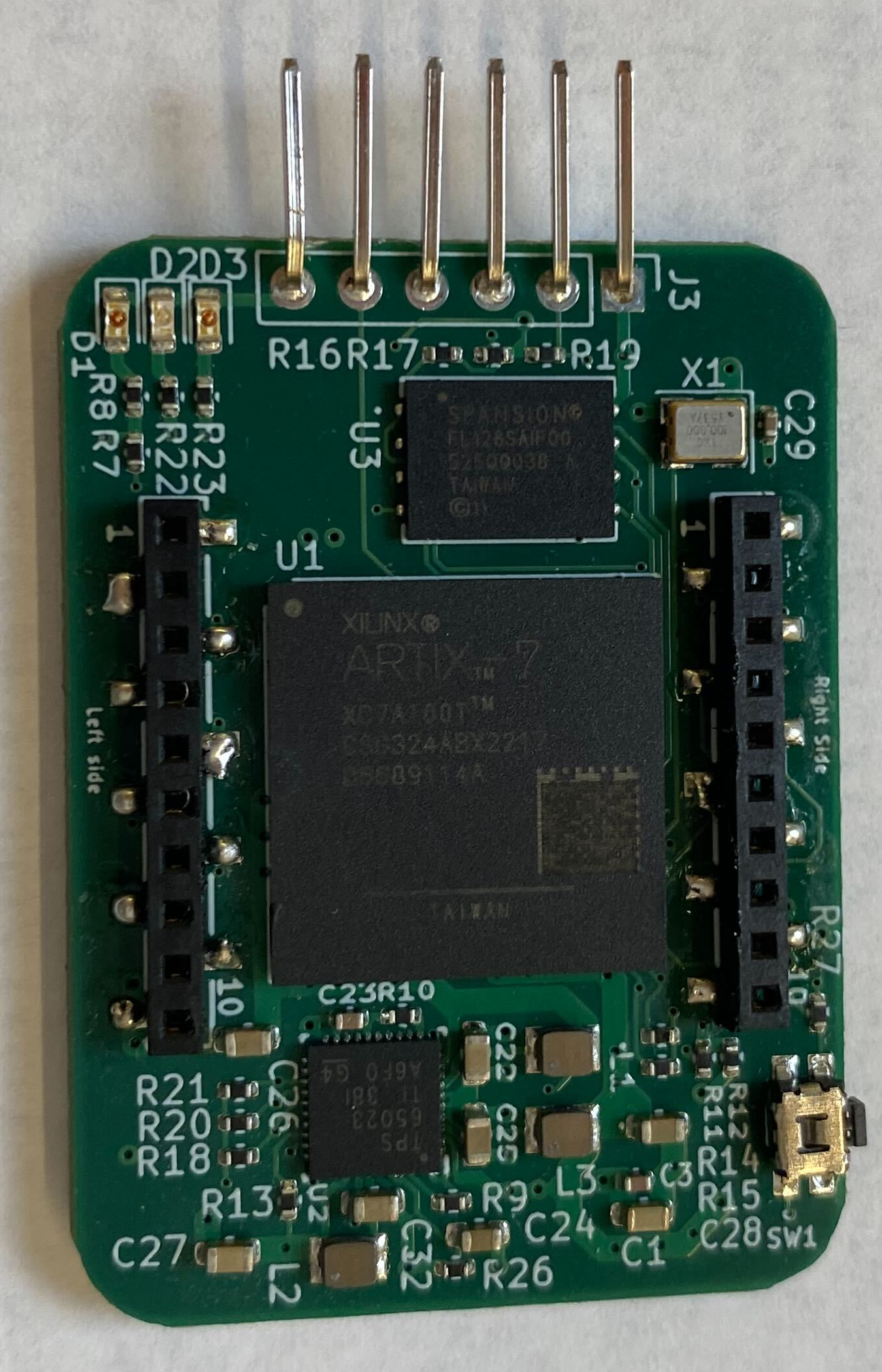}
  \end{minipage}\hfill
  \begin{minipage}{0.49\linewidth}
    \centering
    \includegraphics[width=\linewidth]{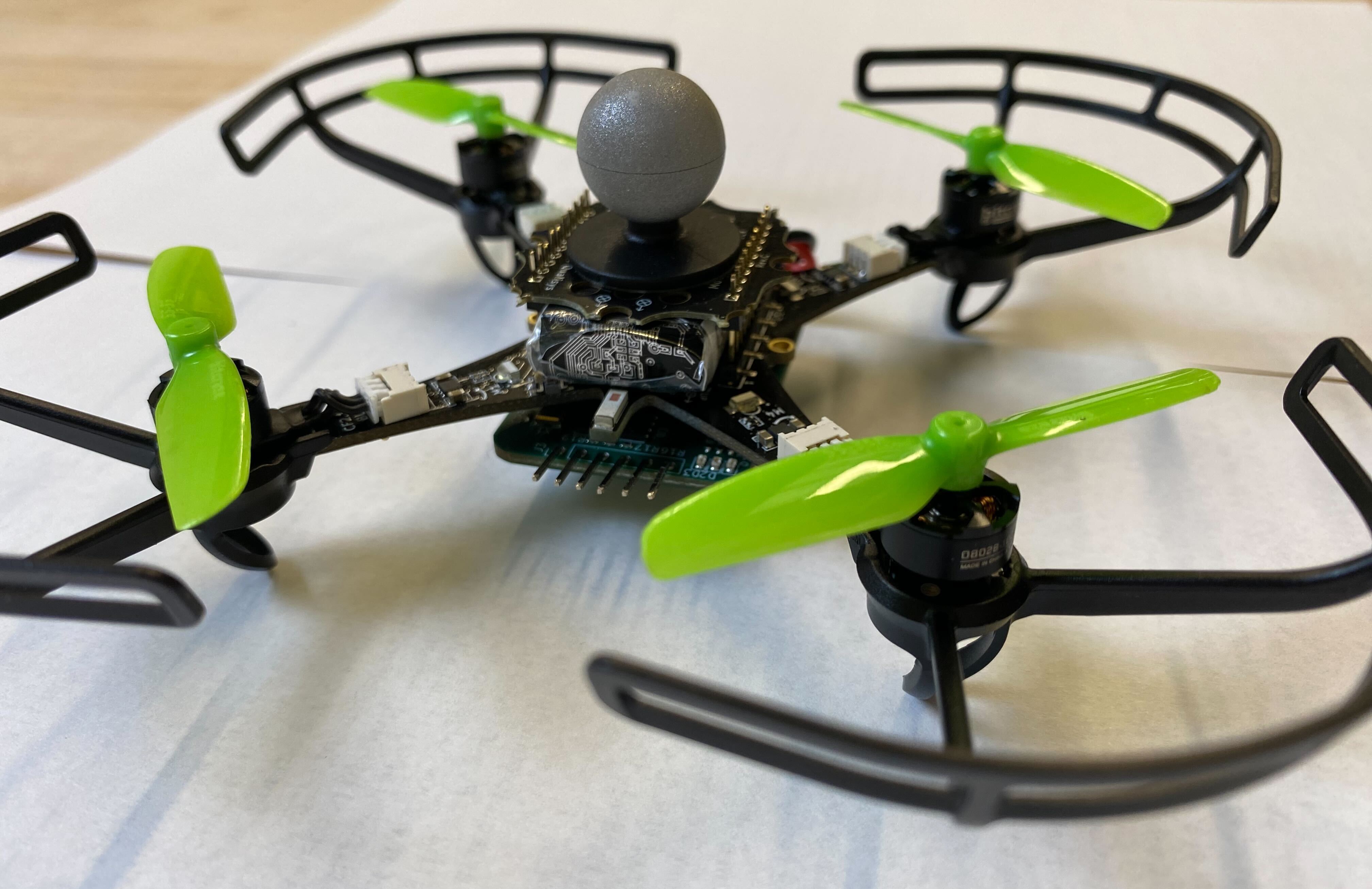}
  \end{minipage}
  \caption{The 4$\times$\SI{3}{\cm} four-layer PCB hosts an AMD Artix-7~100T FPGA and mounts directly on the Crazyflie through the standard expansion interface. Left: fabricated FPGA board. Right: fully integrated onboard the tiny quadrotor.}
  \label{fig:pcb}
  \vspace{-10pt}
\end{figure}

We found that existing solutions to mount an FPGA on the Crazyflie were unsuitable for kilohertz-rate onboard optimization. Available decks either provide insufficient FPGA resources (e.g.,~\cite{bitcraze_lighthouse_deck_fpga}) or rely on low-bandwidth interfaces such as I\textsuperscript{2}C (e.g.,~\cite{azemFPGABasedNeuralThrust2024}), which impose excessive communication latency. We therefore designed and fabricated a custom four-layer PCB expansion board (Fig.~\ref{fig:pcb}).

The board hosts an AMD Artix-7~100T FPGA powered by a dedicated PMIC that generates and sequences the required core and I/O voltage rails. A high-speed SPI interface connects the FPGA directly to the Crazyflie MCU, enabling low-latency exchange of the current optimization data and the resulting control input. A \SI{100}{\mega\hertz} oscillator provides the solver clock, while an onboard SPI flash stores the FPGA configuration for autonomous startup. Additional JTAG access and user LEDs support programming and debugging.

To enable the integration of our custom PCB, we modified the FreeRTOS-based Crazyflie firmware to compute the deviation between the EKF state estimate and the current reference, transmit the runtime solver data to the FPGA, trigger the optimization, and retrieve the first optimal control input for actuation. Runtime constraint bounds can be updated through the same interface, allowing the admissible set to change from one control cycle to the next and supporting dynamic obstacle-avoidance (e.g., Sec.~\ref{subsec:flight_experiments}). As a result, the complete ADMM solve is offloaded to the FPGA, while state estimation, communication, safety logic, and the remaining flight stack remain on the MCU. This division allows the accelerator to be integrated without replacing the established Crazyflie autonomy stack.

\section{Experimental Evaluation} \label{sec:exp_ev}
\subsection{Methodology}
We evaluate the proposed solver in three complementary settings.
First, we benchmark TinyMPC on the Crazyflie MCU (STM32F405, 168\,MHz) with the FPGA implementations over prediction horizons $H \in \{10,20,30,40,50,60,70,80\}$.
TinyMPC is measured inside the FreeRTOS flight stack over 80 samples per operating point, whereas FPGA execution is cycle-deterministic and therefore requires only a single measurement per synthesized configuration.
Second, we compare in more detail the two FPGA mappings, \texttt{full\_sparse} and \texttt{staged}, over the range where both implementations fit on the target Artix-7 FPGA and discuss the scalability of the \texttt{staged} mapping, extending the prediction horizon to $H=1350$.
Finally, we deploy our approach onto a Crazyflie Brushless~2.1 in an OptiTrack eight-camera arena with a 120\,Hz position stream. The ADMM control loop is fully offloaded to the onboard FPGA and runs at \SI{1}{\kilo\hertz}. During flight, the remaining standard Crazyflie flight stack executes on the onboard MCU.

\subsection{Solver Benchmarks}
\begin{figure*}[!t]
    \centering
    \includegraphics[width=\linewidth]{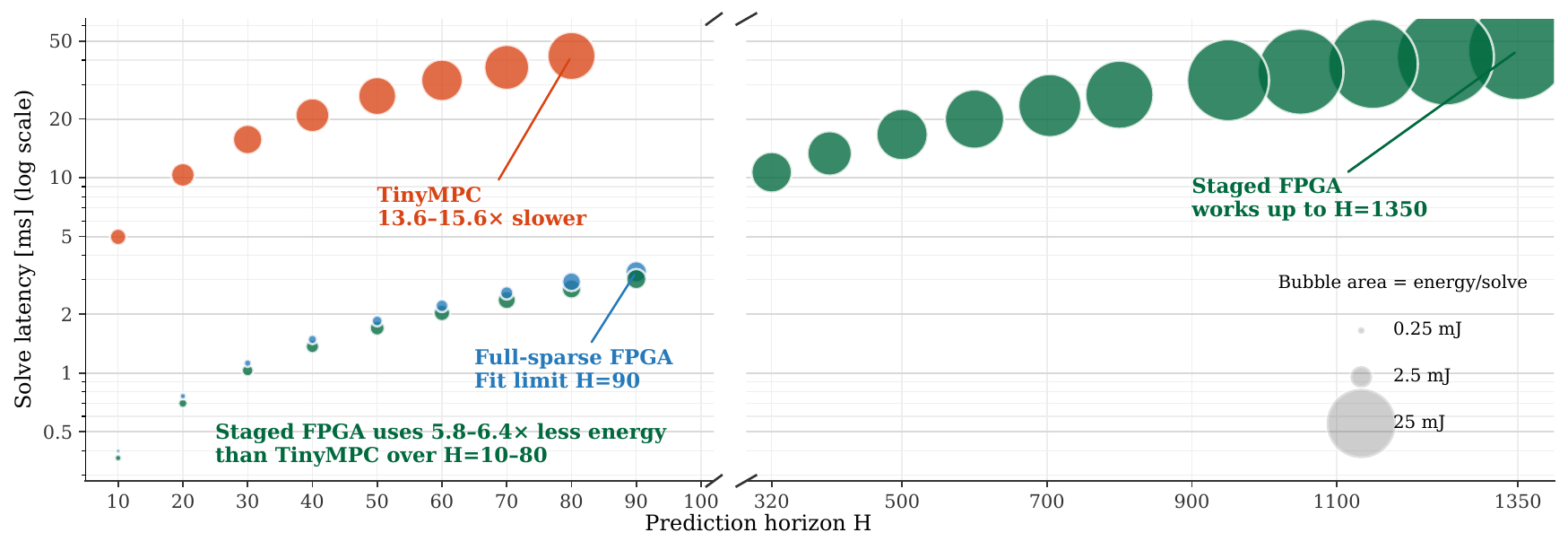}
    \caption{Latency and energy versus prediction horizon for the \texttt{staged} FPGA design, \texttt{full\_sparse} FPGA baseline, and TinyMPC at a fixed ADMM budget of k=10. Marker area is proportional to energy per solve.}
    \label{fig:latencyEnergy}
\end{figure*}

\begin{figure}[t]
  \centering
  \includegraphics[width=\linewidth]{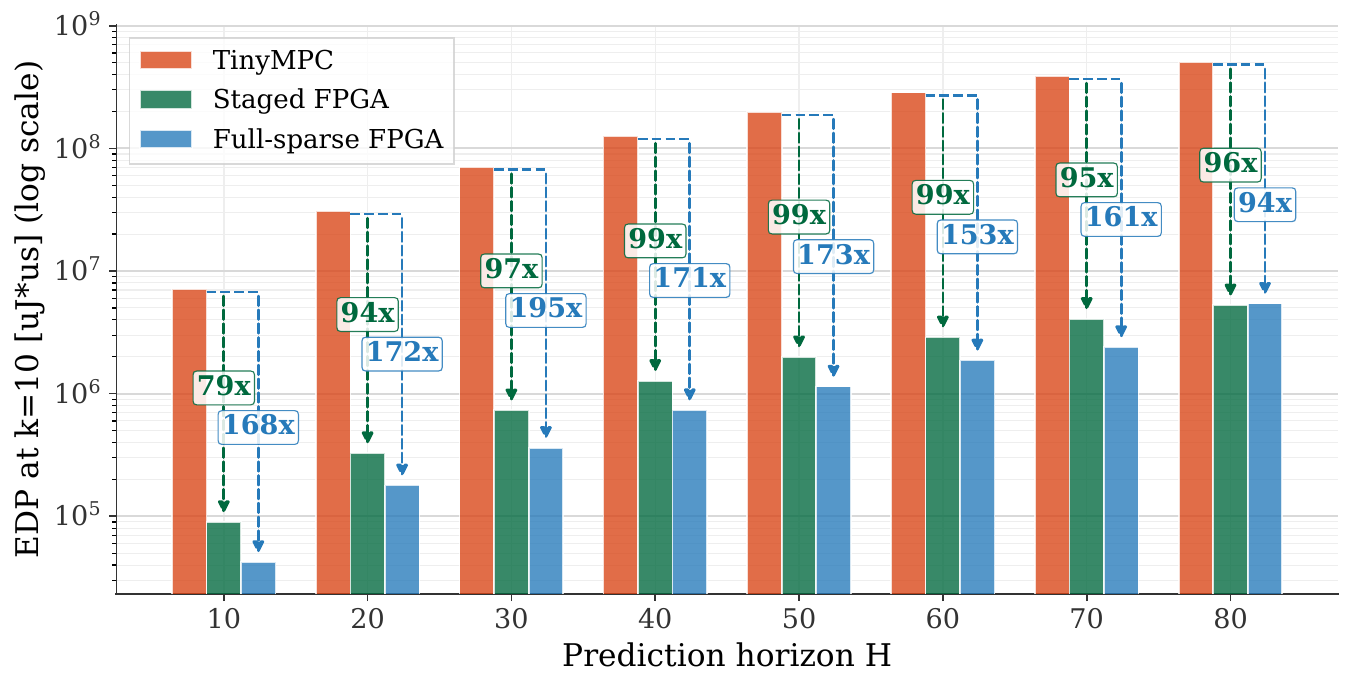}
  \caption{Energy-delay product at $10$ ADMM iterations across various horizons. FPGA implementations achieve up to 195.4$\times$ lower EDP than TinyMPC.}
  \label{fig:edp}
\end{figure}

Fig.~\ref{fig:latencyEnergy} shows the solve time versus prediction horizon, along with the energy per solve, for TinyMPC and both the \texttt{full\_sparse} and \texttt{staged} FPGA implementations.
While all three implementations scale approximately linearly with horizon, the FPGA implementations are much faster and consume far less energy.
In particular, the FPGA implementations are up to $15.6\times$ faster than TinyMPC. The median time per ADMM iteration per horizon step is 52.3\,\si{\micro s}/(iter$\cdot$step) for TinyMPC, 3.69\,\si{\micro s}/(iter$\cdot$step) for \texttt{full\_sparse}, and 3.33\,\si{\micro s}/(iter$\cdot$step) for \texttt{staged}.
Similarly, the FPGA implementations achieve 5.8-14.0$\times$ lower energy per solve and 79.1-195.4$\times$ lower energy-delay product (EDP) than TinyMPC (Fig.~\ref{fig:edp}). As expected, \texttt{full\_sparse} uses far less power but is unable to scale as well as \texttt{staged}. 
For TinyMPC, MCU power consumption is evaluated under active operational parameters ($87\text{ mA}$ at $168\text{ MHz}$, $3.3\text{ V}$), establishing a nominal compute power profile for the STM32F405. FPGA power is obtained from post-route power analysis combined with the measured solve time.

Table~\ref{tab:feasibility} also reports the maximum feasible ADMM iteration budget within the target \SI{1}{\kilo\hertz} control period, including MCU-FPGA communication overhead.
We find that within a \SI{1}{\kilo\hertz} control period, TinyMPC cannot sustain a sufficient optimization budget for constrained MPC even at the shortest prediction horizons, and quickly becomes unable to complete a single ADMM iteration as the horizon increases. The FPGA implementations instead preserve multiple iterations across the evaluated horizons, making kilohertz-rate constrained MPC practical on the target platform.

\begin{table*}[t]
\centering
\caption{Maximum feasible ADMM iterations within a \SI{1}{\kilo\hertz} control period, FPGA includes 100\,\si{\micro s} SPI overhead.}
\begin{tabular}{lccccccccccccc}
\toprule
& \multicolumn{13}{c}{Prediction Horizon $H$} \\
\cmidrule(lr){2-14}
Solver & 5 & 10 & 20 & 30 & 40 & 50 & 70 & 90 & 100 & 150 & 200 & 250 & 275 \\
\midrule
TinyMPC (MCU) & 4 & 2 & -- & -- & -- & -- & -- & -- & -- & -- & -- & -- & -- \\
\texttt{Full\_sparse} FPGA & 41 & 22 & 11 & \textbf{8} & \textbf{6} & 4 & \textbf{3} & \textbf{2} & -- & -- & -- & -- & -- \\
\texttt{Staged} FPGA & \textbf{43} & \textbf{24} & \textbf{12} & \textbf{8} & \textbf{6} & \textbf{5} & \textbf{3} & \textbf{2} & \textbf{2} & \textbf{1} & \textbf{1} & \textbf{1} & -- \\
\bottomrule
\end{tabular}
\label{tab:feasibility}
\end{table*}

\subsection{FPGA Architecture Tradeoffs and Scalability}
\begin{figure}[!t]
  \centering
  \includegraphics[width=0.95\linewidth]{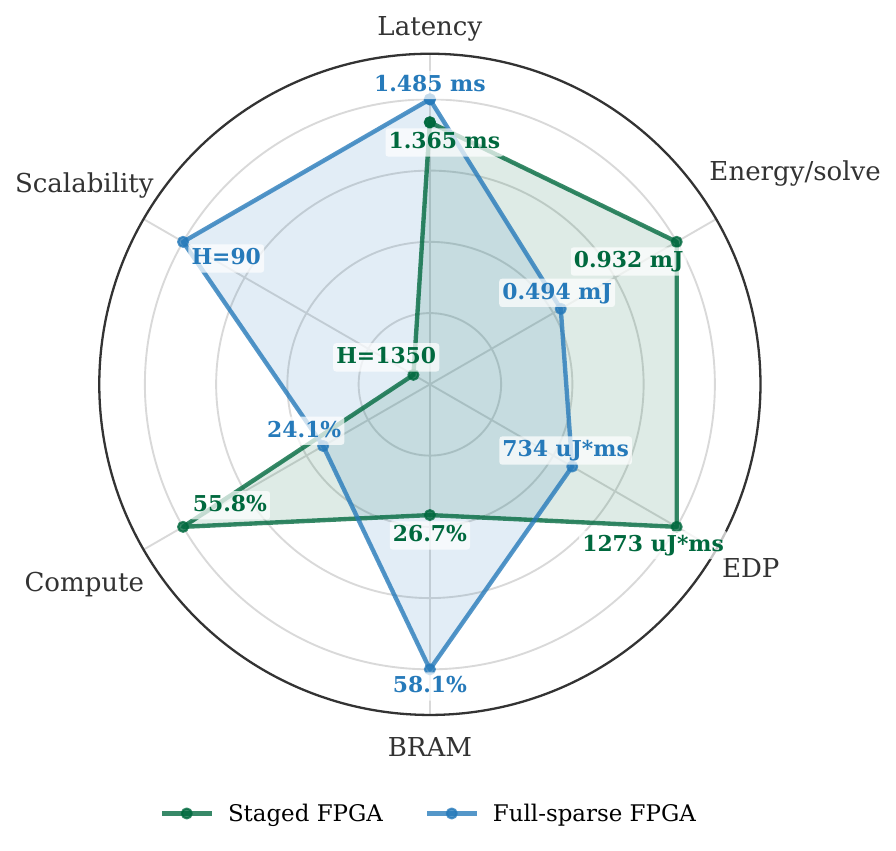}
  \caption{A multi-objective comparison of the \texttt{staged} and \texttt{full\_sparse} FPGA solvers at $H{=}40$ and $k{=}10$ ADMM iterations. All metrics are scaled so that lower is better.}
  \label{fig:solv_comparison}
  \vspace{-10pt}
\end{figure}

We next compare the \texttt{full\_sparse} and \texttt{staged} FPGA mappings in additional detail. Overall, as shown in Figure~\ref{fig:solv_comparison}, neither mapping dominates across all design objectives and both provide high performance and low power control.

As noted above, Fig.~\ref{fig:latencyEnergy} shows that both implementations have nearly identical latency scaling over their shared operating range, with the \texttt{staged} mapping consistently providing slightly lower latency.
In contrast, \texttt{full\_sparse} achieves lower power and energy per solve in the horizon range where its expanded operators fit efficiently in on-chip memory.

These performance differences reflect the underlying resource usage.
The \texttt{full\_sparse} mapping prioritizes specialized datapaths and lower compute utilization at the expense of increasing BRAM and LUTRAM pressure as the horizon grows.
Conversely, \texttt{staged} trades additional computation for a substantially more compact memory footprint, allowing the implementation to avoid the LUTRAM spillover that limits \texttt{full\_sparse} near $H=90$. This compact footprint allows \texttt{staged} to scale to $H=1350$, corresponding to 21,612 optimization variables and 24,312 constraints, while maintaining a solve time of \SI{44.9}{\milli\second} for $k=10$ ADMM iterations, equivalent to 22.3 online solves per second. Although this operating point targets slower robotic control or planning loops rather than kilohertz-rate quadrotor control, it demonstrates that the same solver architecture extends well beyond the embedded flight regime without algorithmic modifications.

\subsection{Onboard Flight Experiments}
\label{subsec:flight_experiments}

\subsubsection{Unconstrained Figure-Eight Tracking}
\begin{figure}[t]
  \centering
  \begin{minipage}{0.57\linewidth}
    \centering
    \includegraphics[width=\linewidth]
    {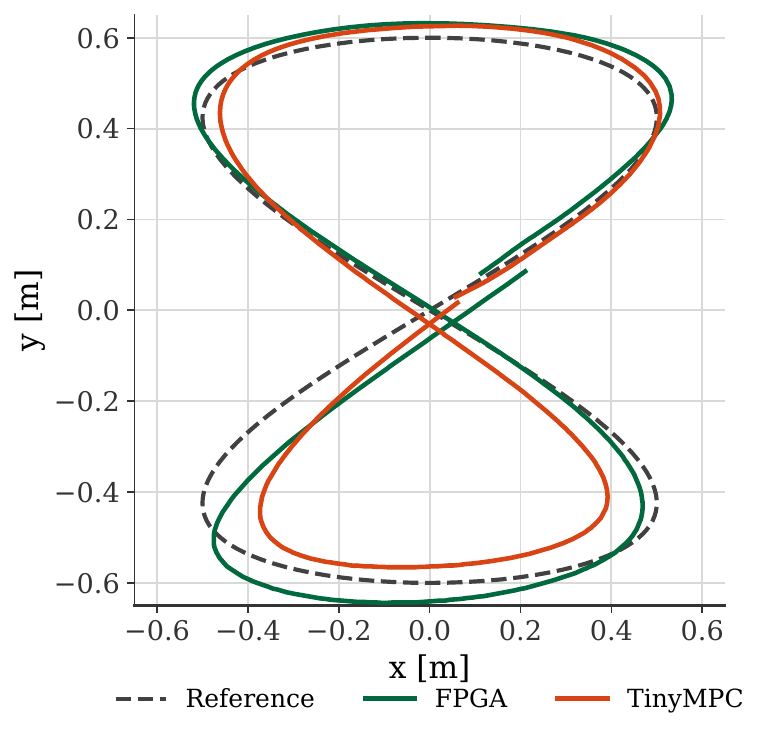}
  \end{minipage}\hfill
  \begin{minipage}{0.38\linewidth}
    \centering
    \includegraphics[width=\linewidth]{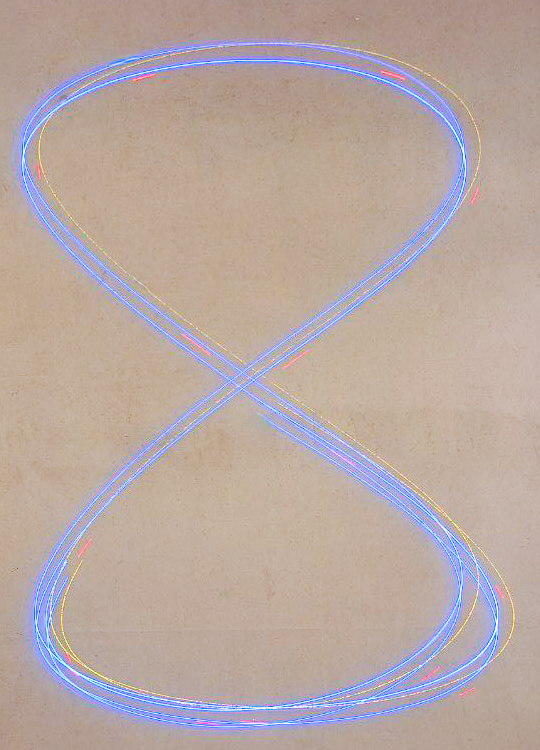}
  \end{minipage}
  \caption{Unconstrained figure-eight trajectory tracking. TinyMPC and the FPGA-based controller achieve comparable closed-loop tracking performance. Left: trajectory overlay in the $xy$ plane. Right: long-exposure flight photograph.}
  \label{fig:figure_eight}
  \vspace{-10pt}
\end{figure}

We evaluate nominal tracking performance on a smooth figure-eight trajectory in the horizontal plane at constant altitude.
TinyMPC uses $H{=}15$ and $k{=}2$ at 500~Hz, while the FPGA controller is able to scale to $H{=}20$ and $k{=}28$ due to its faster performance within the same control rate.
As no constraints are active in this experiment, both configurations solve tractable QPs and achieve comparable closed-loop tracking performance, as shown in Fig.~\ref{fig:figure_eight}.
This experiment verifies that the FPGA implementation preserves the nominal tracking behavior of the embedded baseline while providing a substantially larger online optimization budget.

\subsubsection{Static constrained trajectory tracking}
We restrict the admissible workspace to a square region via box constraints on the $x$ and $y$ coordinates, and design a star-shaped reference that intentionally violates these bounds. This forces the controller to modify the commanded motion online to maintain feasibility.

\begin{figure}[t]
  \centering
  \includegraphics[width=\linewidth]{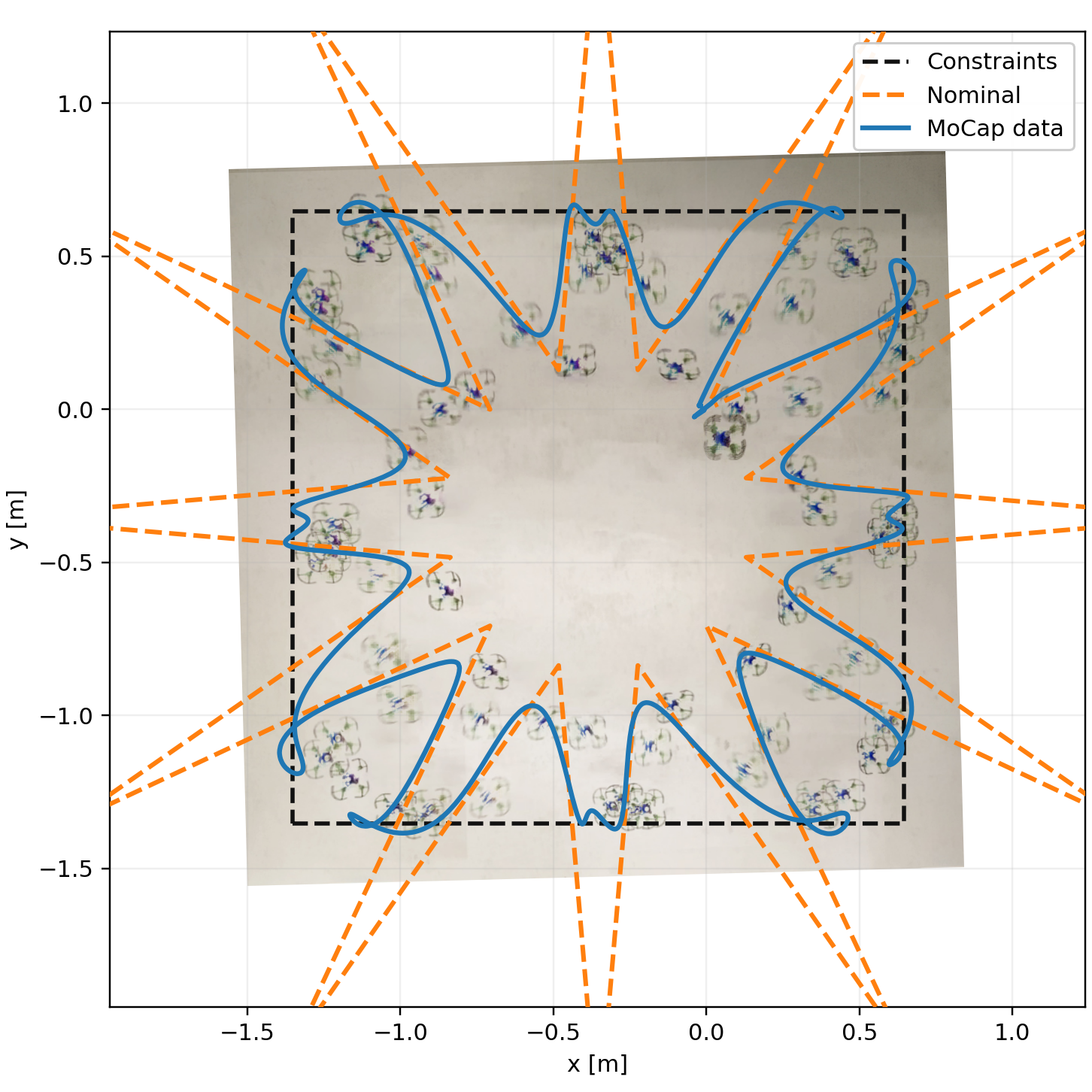}
  \vspace{-15pt}
  \caption{Hardware demonstration of constrained trajectory tracking at 1~kHz. Box constraints on $x$ and $y$ are enforced fully onboard by the FPGA-accelerated MPC controller ($H=27, k=9$). The trajectory remains within the admissible region despite a constraint-violating reference.}
  \label{fig:constrained_star}
  \vspace{-10pt}
\end{figure}

Fig.~\ref{fig:constrained_star} shows that the measured trajectory remains inside the admissible region throughout the physical flight.
At \SI{1}{\kilo\hertz}, TinyMPC cannot execute enough ADMM iterations to reliably converge to a feasible constrained solution, even at short horizons.
The FPGA implementation instead solves the constrained problem with $H{=}27$ and $k{=}9$ within every control cycle, enabling kilohertz-rate constraint enforcement onboard a tiny quadrotor.

\subsubsection{Dynamic obstacle avoidance}
Finally, we evaluate the ability of the controller to react online to dynamically moving obstacles without modifying the FPGA configuration. The quadrotor tracks a nominal figure-eight reference while two operators move instrumented boxes through the flight volume and across the nominal trajectory.

Each box is tracked as an OptiTrack rigid body, and its measured position is transmitted to the quadrotor at the motion-capture update rate of \SI{120}{\hertz}. Each obstacle is represented by a separating half-space with a fixed normal and an offset determined from the measured obstacle position and a prescribed safety margin. As such, obstacle motion changes only the corresponding entries of $l$ and $u$, while the constraint matrix remains fixed. Thus, the precomputed Cholesky factorization and the FPGA bitstream remain unchanged throughout the experiment, allowing the onboard controller to incorporate these updated bounds while continuing to solve the MPC problem at \SI{1}{\kilo\hertz}.

As shown in Fig.~\ref{fig:dynObsAvoid}, the quadrotor follows the nominal figure-eight trajectory while the path is unobstructed. When either box moves into the nominal flight path, the obstacle-dependent constraints modify the admissible region, causing the optimized trajectory to deviate from the reference and avoid the obstacle. Once the path becomes clear, the quadrotor returns to the nominal trajectory.


\section{Conclusion and Future Work} \label{sec:conclusion}
We presented AccelMPC, an open source, FPGA-accelerated linear MPC controller paired with a custom PCB for fast, low-latency communication. AccelMPC achieves up to a $15.6\times$ speedup and $195.4\times$ improvement in energy-delay product over state-of-the-art embedded MCU solvers, enables \SI{1}{\kilo\hertz} constrained trajectory tracking on a tiny quadrotor, and scales to problems with over 20{,}000 optimization variables on the same solver architecture. 

Future work will extend the system to richer collision-avoidance formulations and more aggressive flight regimes. Cluttered and fully three-dimensional environments, online replanning around moving obstacles, and drone-racing-style trajectories introduce more complex geometric and dynamic constraints while pushing the vehicle closer to its actuation limits~\cite{kaufmann2023champion,hanover2024drone,kamel2017robust,romero2022replanning,krinner2024mpccpp,penicka2022minimum}. These settings provide a natural testbed for evaluating how high-rate hardware-accelerated control can support more sophisticated models and constraints under the stringent compute and power budgets of tiny aerial robots.
Future work will also more systematically analyze the numerical effects of reduced-precision arithmetic, for example using automated precision-analysis tools such as RoboPrec~\cite{yilmaz2026roboprec}, with the goal of establishing accuracy and constraint-satisfaction guarantees for the current fixed-point implementation. Building on these guarantees, more aggressive fixed-point representations could then be explored to further reduce FPGA resource usage and power while preserving closed-loop control performance.

\bibliographystyle{styles/IEEEtran_new}
\bibliography{styles/IEEEabrv,a2r}

\end{document}